\pdfoutput=1
\documentclass[11pt]{article}

\usepackage[pdftex]{graphicx}
\usepackage{amsmath,mathtools}
\usepackage{amssymb}
\usepackage{booktabs}
\usepackage{makecell}

\usepackage{xcolor}
\usepackage{color}
\usepackage{colortbl}
\definecolor{mygray}{gray}{.9}

\usepackage{ACL2023}

\usepackage{times}
\usepackage{latexsym}

\usepackage[T1]{fontenc}
\usepackage[utf8]{inputenc}

\usepackage{microtype}

\usepackage{inconsolata}
\usepackage{adjustbox}

\title{AD-KD: Attribution-Driven Knowledge Distillation for\\  Language Model Compression}

\author{Siyue Wu\textsuperscript{1}, Hongzhan Chen\textsuperscript{1}, Xiaojun Quan\textsuperscript{1}\thanks{\; Corresponding author.}, Qifan Wang\textsuperscript{2} and Rui Wang\textsuperscript{3} \\
  \textsuperscript{1}School of Computer Science and Engineering, Sun Yat-sen University, China \\
  \textsuperscript{2}Meta AI \\
  \textsuperscript{3}Vipshop (China) Co., Ltd., China \\
  \textsuperscript{1}\texttt{ \{wusy39, chenhzh59\}@mail2.sysu.edu.cn, quanxj3@mail.sysu.edu.cn}\\
  \textsuperscript{2}\texttt{wqfcr@fb.com} \\
  \textsuperscript{3}\texttt{mars198356@hotmail.com}} 

\begin{document}
\maketitle
\begin{abstract}
%This document is a supplement to the general instructions for *ACL authors. It contains instructions for using the \LaTeX{} style file for ACL 2023.
%The document itself conforms to its own specifications, and is, therefore, an example of what your manuscript should look like.
%These instructions should be used both for papers submitted for review and for final versions of accepted papers.

Knowledge distillation has attracted a great deal of interest recently to compress pre-trained language models. However, existing knowledge distillation methods suffer from two limitations. First, the student model simply imitates the teacher's behavior while ignoring the underlying reasoning. Second, these methods usually focus on the transfer of sophisticated model-specific knowledge but overlook data-specific knowledge. In this paper, we present a novel attribution-driven knowledge distillation approach, which explores the token-level rationale behind the teacher model based on Integrated Gradients (IG) and transfers attribution knowledge to the student model. To enhance the knowledge transfer of model reasoning and generalization, we further explore multi-view attribution distillation on all potential decisions of the teacher. Comprehensive experiments are conducted with BERT on the GLUE benchmark. The experimental results demonstrate the superior performance of our approach to several state-of-the-art methods. 
\end{abstract}

\section{Introduction}

Transformer-based pre-trained language models (PLMs), such as BERT \citep{devlin-etal-2019-bert} and RoBERTa \citep{liu2019roberta}, have aroused widespread interest among Natural Language Processing (NLP) researchers in recent years. These language models are first pre-trained on large-scale unlabeled corpora to learn the general representation of language, and then fine-tuned on specific downstream tasks to effectively transfer the general knowledge to target domains. This pre-training and fine-tuning paradigm leads to state-of-the-art performances in various NLP tasks such as natural language understanding. However, with the rapid growth of the model scale, the deployment of large-scale PLMs becomes challenging, especially in low-resource scenarios. To this end, a variety of model compression techniques have been developed. Among them, knowledge distillation (KD) \citep{hinton2015distilling} is a newly emerging technology that aims to obtain a small student model by distilling knowledge from a large teacher model and achieve comparable performance.

\begin{figure}[t]
	\centering
        \includegraphics[scale=0.42]{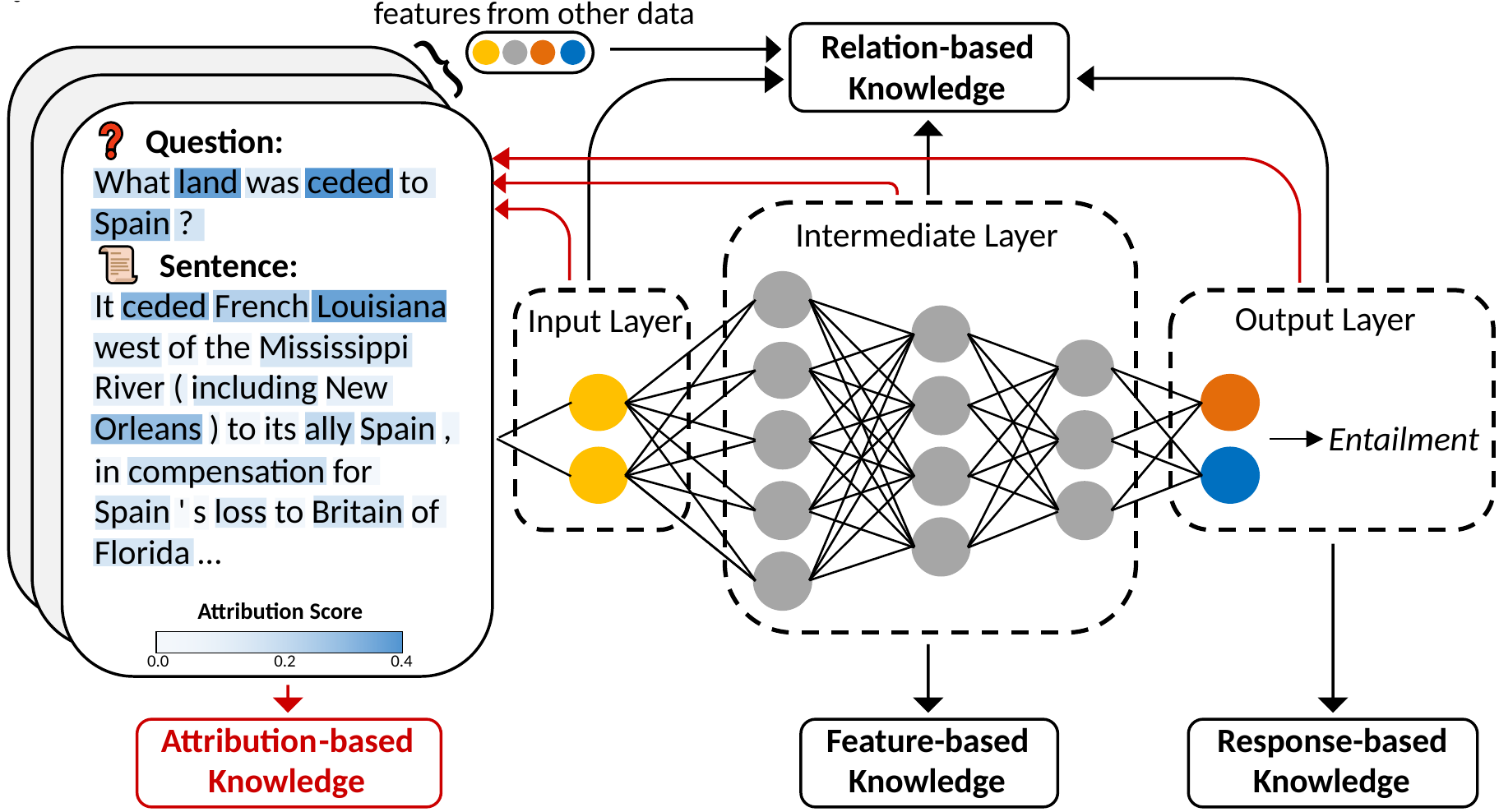}
	\caption{An example from the QNLI dataset \citep{rajpurkar-etal-2016-squad} to illustrate different knowledge distillation techniques including the proposed attribution-driven method. Darker colors mean larger attribution scores.}
	\label{fig:example}
 \vspace{-5mm}
\end{figure}

%In general, KD can be divided into two categories, namely logit-based KD and intermediate feature-based KD. While logit-based KD directly learns final output like probability distribution from the top of the teacher, intermediate feature-based KD makes alignment to match the feature from intermediate layers between teacher and student. To provide comprehensive supervision from top to bottom, a common practice is to combine both logit-based KD and intermediate feature-based KD. However, empirically intermediate feature-based KD brings limited improvement, and sometimes is even inferior to logit-based KD (i.e., vanilla KD).  Some works \citep{liang2022less} find that it is the capacity gap between teacher and student that brings difficulty to student in mimicking the teacher at every layer. In this paper, we go a step further and attribute the aforementioned phenomenon to two reasons. Firstly, existing features mainly focus on what teacher behaves like/what the teacher's behaviors are instead of why teacher behaves like this/why teacher has such behaviors, hindering the generalization ability of student. Secondly, existing features mainly reflect model-specific knowledge but fail to capture the data-specific knowledge which is more general and easier to transfer between different models.

%In terms of different forms of knowledge, 
Existing knowledge distillation methods can be divided into three categories, namely response-based, feature-based, and relation-based \cite{gou2021knowledge}. While response-based methods \cite{turc2019well} directly distill the final output, e.g. probability distribution, from the top of the teacher, feature-based \cite{sun-etal-2019-patient} and relation-based methods \cite{liu-etal-2022-multi-granularity} try to align the features from intermediate layers of teacher and student models and minimize the difference. To transfer comprehensive knowledge from the teacher, a common practice is to combine response-based methods with the other two \cite{park-etal-2021-distilling}. However, due to the capacity gap between the teacher and the student, feature-based and relation-based methods may not necessarily bring improvement to response-based methods \citep{liang2022less}. To sum up, existing knowledge distillation methods have two limitations. First, they mainly focus on understanding what the teacher's behavior is, instead of why the teacher behaves like this, hindering the reasoning and generalization ability of the student model. Second, they pay more attention to distilling sophisticated model-specific knowledge from intermediate layers but neglect data-specific knowledge, which may contain valuable rationale information to understand how the teacher model arrives at a prediction.

%Inspired by the research of attribution, in this paper we propose to transfer a novel kind of feature, namely attribution maps of input tokens. As shown in Figure \ref{fig:example}, the attribution maps highlight the keywords towards the label and can be seen as a kind of knowledge which is complementary to the soft label knowledge. By transferring such attribution knowledge, the student is able to learn the token-level rationale behind teacher's behaviors instead of the behaviors itself and thus generalizes better. Specifically, we utilize Integrated Gradients (IG) \citep{sundararajan2017axiomatic}, a well-established gradient-based attribution method, to calculate the importance score of each input token. To alleviate the interference of noisy dimension in the input embedding of teacher, we further leverage top-\textit{K} method to filter out dimensions with low attribution score, while the remaining ones with high attribution score are reduced to produce a scalar value representing the importance of individual token. Moreover, we extract the attribution knowledge from all potential decisions of model rather than just the decision with maximum probability. By transferring this kind of multi-view attribution knowledge, student can get a more comprehensive understanding of teacher's soft label distribution.

To address the above limitations, in this paper we propose a novel Attribution-Driven Knowledge Distillation (AD-KD) approach that transfers attribution-based knowledge from the teacher to the student.
%, namely attribution maps of input tokens. 
As shown in Figure \ref{fig:example}, the attribution information reflects the importance of different tokens towards the prediction, which contains reasoning knowledge of the model and can be complementary to the soft-label knowledge. By transferring such attribution knowledge, the student is allowed to learn the token-level rationale behind the teacher's behavior and thus generalizes better. Specifically, we utilize Integrated Gradients (IG) \citep{sundararajan2017axiomatic}, a well-established gradient-based attribution method, to calculate the importance score of each input token. To reduce the influence of trivial dimensions in the teacher's input embeddings, we further adopt the top-\textit{K} strategy to filter out dimensions with low attribution scores. The remaining attribution scores are aggregated and normalized to denote the importance of individual tokens. Moreover, we extract the attribution knowledge for all possible predictions rather than just the prediction with the highest probability. By transferring the multi-view attribution knowledge, the student learns a more comprehensive understanding of the teacher's soft-label distribution.

%Extensive experiments are conducted with BERT \citep{devlin-etal-2019-bert} on General Language Understanding Evaluation (GLUE) benchmark \citep{wang-etal-2018-glue} and experimental results demonstrate the effectiveness and superiority of our method over other baselines. Furthermore, we also show that the attribution knowledge behind input layer is indeed more helpful than the attribution knowledge behind intermediate layers in the context of knowledge distillation. To summarize, the contributions of this paper are threefold. First, we propose Attribution-based Knowledge Distillation (Attr-KD) framework for language model compression that transfers a novel kind of feature, namely attribution maps of input token, between teacher and student. Second, we extract multi-view attribution knowledge based on all potential decisions of model instead of single-view attribution knowledge to teach the student comprehensively. Third, we validate Attr-KD on GLUE benchmark and show that Attr-KD achieves competitive performance compared to other knowledge distillation baselines.

Extensive experiments are conducted with BERT \citep{devlin-etal-2019-bert} on the GLUE benchmark \citep{wang-etal-2018-glue}.~The experimental results demonstrate the effectiveness and superiority of our approach over several state-of-the-art baselines. Furthermore, we show that attribution knowledge from different layers contains different information, while the input layer contains the most prominent attribution knowledge for distillation. To summarize, the main contributions are threefold. First, we propose a novel attribution-driven knowledge distillation framework for language model compression that effectively transfers attribution knowledge from the teacher to the student. Second, we extract multi-view attribution knowledge based on model predictions to learn comprehensive reasoning knowledge. Third, we systematically validate AD-KD on the GLUE benchmark and show its superior performance over state-of-the-art baselines.

%These instructions are for authors submitting papers to ACL 2023 using \LaTeX. They are not self-contained. All authors must follow the general instructions for *ACL proceedings,\footnote{\url{http://acl-org.github.io/ACLPUB/formatting.html}} as well as guidelines set forth in the ACL 2023 call for papers.\footnote{\url{https://2023.aclweb.org/calls/main_conference/}} This document contains additional instructions for the \LaTeX{} style files.
%The templates include the \LaTeX{} source of this document (\texttt{acl2023.tex}),
%the \LaTeX{} style file used to format it (\texttt{acl2023.sty}),
%an ACL bibliography style (\texttt{acl\_natbib.bst}),
%an example bibliography (\texttt{custom.bib}),
%and the bibliography for the ACL Anthology (\texttt{anthology.bib}).
%\vspace{-2mm}
\section{Related Work}
\subsection{Knowledge Distillation}
%Logit-based KD was first proposed by \citet{hinton2015distilling}, where the final output is adopted to transfer the label knowledge. Specifically, they minimized the KL-divergence between the probability distributions of teacher and student. \citet{sanh2019distilbert} and \citet{turc2019well} applied this idea to BERT and yielded smaller models with minor performance drop. 
Knowledge distillation methods can be divided into three categories, namely response-based, feature-based and relation-based KD \citep{gou2021knowledge}.
Response-based KD was first proposed by \citet{hinton2015distilling}, where the final output is adopted to transfer the label knowledge. 
%Specifically, they minimized the KL-divergence between the probability distributions of teacher and student.
\citet{sanh2019distilbert} and \citet{turc2019well} applied this idea to BERT and yielded smaller models with minor performance drops. 
Recently, feature-based and relation-based distillation methods have drawn a lot of attention, which transfer knowledge contained in the intermediate layers to the student. For feature-based methods, \citet{sun-etal-2019-patient} first regarded the hidden representations of the [CLS] token as hints to extract sentence-level features from the teacher. \citet{jiao-etal-2020-tinybert} and \citet{sun-etal-2020-mobilebert} further matched the hidden representations of all tokens between teacher and student models. \citet{sun-etal-2020-contrastive} proposed contrastive distillation on intermediate representations. As for relation-based methods, \citet{park-etal-2021-distilling} proposed CKD which adopts pair-wise distance and triple-wise angle to model the sophisticated relations among token representations from both horizontal and vertical directions. Based on CKD, \citet{liu-etal-2022-multi-granularity} further extracted structural relations from multi-granularity representations and distilled this kind of well-organized multi-granularity structural knowledge hierarchically across layers. \citet{wang2020minilm,wang-etal-2021-minilmv2} generalized the conventional query-key attention to query-query attention, key-key attention, and value-value attention. 
% Although attention matrices can be regarded as one kind of attribution knowledge since they represent the responsibility each input token has on a model prediction to some extent \citep{bastings-filippova-2020-elephant,xu-etal-2020-self}, there are several drawbacks when it comes to distillation. On one hand, attention correlates well with attribution locally in specific layer and head but not globally, indicating that attention maps are inadequate to draw conclusions that refer to the input of the model \citep{pascual-etal-2021-telling}. In other words, attention matrices are more like model-specific knowledge that are probably challenging for student to learn due to the layer mapping issue, especially when the student has much fewer parameters than teacher. On the other hand, some works point out that by adversarial training, alternative attention weights can be found whereas the prediction remains almost the same \citep{jain-wallace-2019-attention,wiegreffe-pinter-2019-attention}. Therefore, an optimal student unnecessarily shares similar attention matrices with its teacher. 
Different from these methods, we investigate knowledge distillation from the attribution perspective, which reveals the teacher's reasoning behavior and can be used to transfer comprehensive data-specific knowledge. More details about the differences between existing methods and ours are discussed in Appendix~\ref{sec:discussion}.

\subsection{Attribution}
Attribution analysis \cite{baehrens2010explain,ancona2018towards} aims at assigning importance scores to intermediate or input features of a network. Occlusion-based methods \citep{zeiler2014visualizing} compute the importance score of each feature by erasing that feature and measuring the difference between new output and the original output. However, occlusion-based methods need to forward pass the model once for each feature, leading to low computational efficiency. To address this issue, gradient-based methods \citep{li-etal-2016-visualizing,ding-etal-2019-saliency,brunner2020identifiability,sundararajan2017axiomatic} exploit the gradient information of features to approximate occlusion-based methods, which only require a single forward process. Similarly, propagation-based methods \citep{bach2015pixel,shrikumar2017learning} modify the back-propagation rules to redistribute the model output among the target features along the back-propagation path.~Perturbation-based methods \citep{guan2019towards,schulz2020restricting,de-cao-etal-2020-decisions} add noise to features to examine their importance for model predictions. 
%The importance score is then estimated by the mutual information between blurred feature and original feature. 
Attribution has been adopted in model compression techniques such as pruning \citep{michel2019sixteen} and adaptive inference \citep{modarressi-etal-2022-adapler} but has not been explored in knowledge distillation. In this work, we take the initiative to investigate the effect of attribution in knowledge distillation.

\begin{figure*}[t]
	\centering
        \includegraphics[scale=0.4]{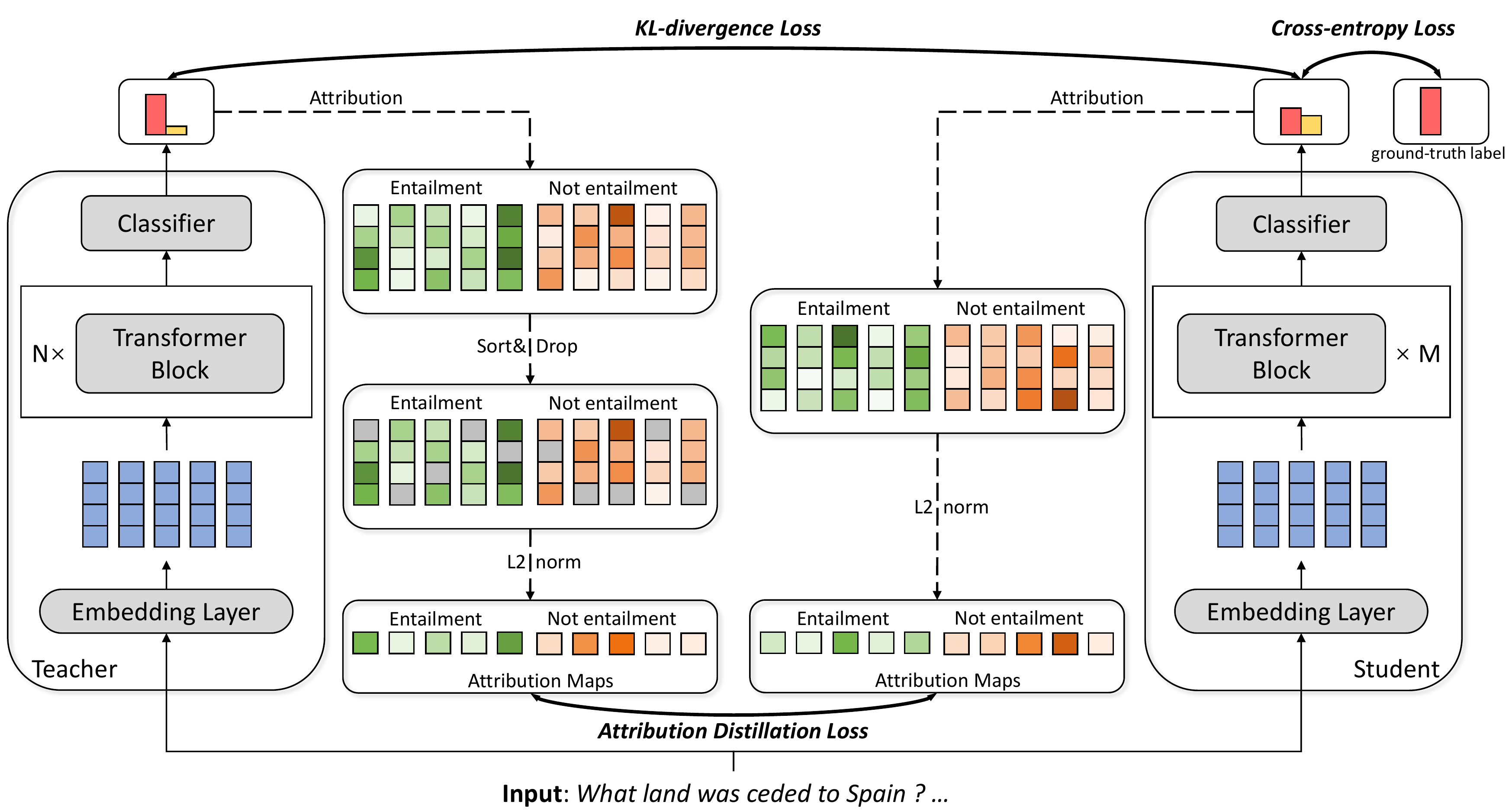}
	\caption{Overview of our AD-KD framework. The example in Figure \ref{fig:example} is taken as the input. AD-KD first extracts the attribution maps from the teacher model and then transfers the attribution-based knowledge to the student.}
	\label{fig:framework}
 \vspace{-4mm}
\end{figure*}

\section{Methodology}

\subsection{Preliminary} \label{sec:background}
Integrated Gradients \citep{sundararajan2017axiomatic} is a theoretically tenable method to attribute the prediction of a deep network to its input or intermediate features. Formally, given a feature $\mathbf{x} = [x_1, x_2, ..., x_n] \in \mathbb{R}^n$ with a baseline feature $\mathbf{x}' = [x'_1, x'_2, ..., x'_n] \in \mathbb{R}^n$, and the model function $F(.)$, IG leverages integral to represent the difference between $F(\mathbf{x})$ and $F(\mathbf{x}')$ by selecting a straight line path from $\mathbf{x}'$ to $\mathbf{x}$ as the integral path: 
\begin{equation}
\small
\begin{aligned}
	F(\mathbf{x})&-F(\mathbf{x}') = \sum_{i=1}^{n} \text{IG}_i (F,\mathbf{x}) =  \\
 \sum_{i=1}^{n}&[(x_i-x'_i)\times \int_{\alpha=0}^1 \frac{\partial F(x'+\alpha \times (x-x'))}{\partial x_i} d\alpha].
\end{aligned}
\end{equation}
% where $\text{IG}$
% Then the integrated gradient of $x_i$ is defined as:
% \begin{equation}
% \begin{aligned}
%     \text{IG}_i&(F,x) = \\
%     (x_i-&x'_i)\times \int_{\alpha=0}^1 \frac{\partial F(x'+\alpha \times (x-x'))}{\partial x_i} d\alpha.
% \end{aligned}
% \end{equation}
In practice, continual integral can be approximated by discrete summation:
\begin{equation}
\small
\begin{aligned}
    \text{IG}_i^{approx}&(F,\mathbf{x}) = \\
    (x_i-x'_i)&\times \sum_{k=1}^m \frac{\partial F(x'+\frac{k}{m} \times (x-x'))}{\partial x_i} \times \frac{1}{m},
\end{aligned}
\end{equation}
where $m$ is the number of summation steps (a bigger $m$ usually results in better approximation). Intuitively, the magnitude of integrated gradient indicates its importance while its sign illustrates the positive or negative effect on the target output.

In this paper, we focus on Transformer-based architecture and attribute the model prediction to input features. With slight abuse of notation, we denote the input sequence as $\mathbf{x} = [x_1, x_2, ..., x_n]$, where $n$ is the sequence length and each $x_i$ represents a token. Transformer first converts the token sequence to $d$-dimensional embedding sequence $\mathbf{E} = [\mathbf{e_1}, \mathbf{e_2}, ... , \mathbf{e_n}] \in \mathbb{R}^{n\times d}$ through the embedding layer. And then the contextualized representations $\mathbf{H} =\text{Transformer}(\mathbf{E}) \in \mathbb{R}^{n\times d}$ are obtained after several layers of Transformer blocks. Finally, a task-specific head is applied on $\mathbf{H}$ to get the final output $P = [P_1, P_2, ... , P_C] \in \mathbb{R}^C$, which is typically a probability distribution. Denote the mapping function $\mathbf{E}\rightarrow P_c$ as $F^c(.)$, where $c$ represents the label of interest. In this case, our attribution map is computed on each individual dimension of each input embedding, which is denoted as $e_{ij}$:
\begin{equation}
\small
\begin{aligned}\label{eq:ig}
    \text{IG}_{ij}^{approx}&(F^c,\mathbf{E}) = \\
    (e_{ij}-e'_{ij})&\times \sum_{k=1}^m \frac{\partial F^c(\mathbf{E}'+\frac{k}{m} \times (\mathbf{E}-\mathbf{E}'))}{\partial e_{ij}} \times \frac{1}{m}.
\end{aligned}
\end{equation}
In the implementation, we stack $n$ \text{[PAD]} token embeddings as baseline features $\mathbf{E}'$ since they usually have no influence on the model prediction.
% , i.e., $\forall ij, \mathbf{E}'_{ij}=0$. Thus Eq. \eqref{eq:ig} is simplified as:
% \begin{equation}
% \small
% \begin{aligned}
%     \text{IG}_{ij}^{approx}(F^c,\mathbf{E}) = 
%     \mathbf{E}_{ij}\times \sum_{k=1}^m \frac{\partial F^c(\frac{k}{m} \times \mathbf{E})}{\partial \mathbf{E}_{ij}} \times \frac{1}{m}
% \end{aligned}
% \end{equation}

\subsection{AD-KD}
In this section, we elaborate on our proposed Attribution-Driven Knowledge Distillation (AD-KD), including attribution maps and attribution distillation. The overall framework of AD-KD is illustrated in Figure \ref{fig:framework}.

\subsubsection{Attribution Maps}
The attribution scores of a language model reflect the importance of different tokens towards the prediction, which contains valuable data-specific reasoning knowledge. 
The scores are computed among different tokens at different dimensions of a given model, using IG defined in Section \ref{sec:background}. In this work, we do not take the sign into consideration, since the scores at different dimensions of the same token embedding would cannibalize each other when combining them into a token-level attribution score. This observation is consistent with the findings in \citep{atanasova-etal-2020-diagnostic}. 

\begin{figure*}[t]
	\centering
        \includegraphics[scale=0.48]{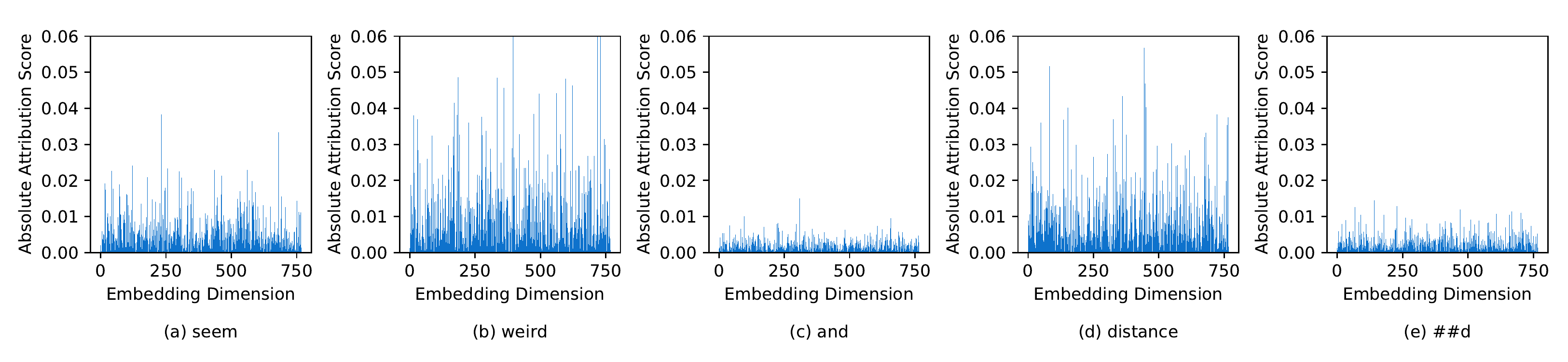}
	\caption{An example from the SST-2 dataset \citep{socher-etal-2013-recursive}. Given the sentence ``seem weird and distanced'' and its sentiment label \emph{negative}, the distributions of absolute attribution scores among different tokens and dimensions are shown in subfigures (a)-(e). The model is a well-trained $\text{BERT}_{base}$ (teacher) and the IG steps $m$ is set to 1.}
	\label{fig:distribution}
 \vspace{-3mm}
\end{figure*}

When calculating the attribution scores, we observed that there exist certain dimensions whose attribution scores remain relatively low across different tokens. The attribution scores from these dimensions minimize the difference between important and unimportant tokens, which can be regarded as noises. For better illustration, Figure~\ref{fig:distribution} shows an example of sentence ``seem weird and distanced'' whose annotation is \emph{negative} sentiment. It is clear that ``weird'' and ``distance'' are the keywords that contribute most to the prediction, whereas a proportion of dimensions of them present low attribution scores. To alleviate the influence of noisy dimensions in the input embeddings, we simply choose the top-\textit{K} dimensions with high attribution scores and filter out dimensions with low attribution scores. Formally, the attribution score of token $x_i$ with respect to the label $c$ in the teacher model can be calculated as:
\begin{equation}
\small
\begin{aligned}
    a_i^{t,c} = \Vert \text{TopK}(\text{IG}_{i}^{approx}(F^{t,c},\mathbf{E}^t))\Vert_2,
\end{aligned}
\end{equation}
where the superscript $t$ denotes the teacher model. Therefore, the attribution map of the teacher consists of a sequence of attribution scores:
\begin{equation}
\small
\begin{aligned}
    \mathbf{a}^{t,c} = [a_1^{t,c},a_2^{t,c},...,a_n^{t,c}].
\end{aligned}
\end{equation}
For the student, the extraction of attribution map is similar except that we consider all dimensions for two reasons. First, it reduces the difficulty of training. Second, the student is allowed to learn from the noiseless attribution map of the teacher.
\begin{equation}
\small
\begin{split}
    a_i^{s,c}=\Vert \text{IG}_{i}^{approx}(F^{s,c},\mathbf{E}^s)\Vert_2, \\
    \mathbf{a}^{s,c}=[a_1^{s,c},a_2^{s,c},...,a_n^{s,c}]. \ \ \
\end{split}
\end{equation}

Considering that the teacher can make multiple decisions, each of which is associated with a probability, we further propose to extract multi-view attribution knowledge. Specifically, we extract the attribution maps for all possible predictions of the model rather than a single prediction, e.g., the prediction with the maximum probability or the prediction corresponding to the ground-truth label. By transferring the multi-view attribution knowledge, the student can capture a more comprehensive understanding of the teacher's soft-label distribution. The multi-view attribution maps are defined as:
\begin{equation}
\small
    \label{eq:multi-view atr maps} \mathbf{A}^{t} = \parallel_{c=1}^{C} \mathbf{a}^{t,c},\ \mathbf{A}^{s} = \parallel_{c=1}^{C} \mathbf{a}^{s,c},
\end{equation}
where $\parallel$ is the concatenation operation.

\subsubsection{Attribution Distillation}
Given the multi-view attribution maps, a straightforward strategy to transfer the knowledge is to directly minimize the difference between the two sets of maps in teacher and student models, with distance metrics like L2 distance (MSE):
\begin{equation}
\small
    \Vert \mathbf{A}^{t} - \mathbf{A}^{s} \Vert_2.
\end{equation}
However, one obvious shortcoming with this approach is that there may exist a magnitude gap between the attribution scores in teacher and student models at the early phase of distillation, since the teacher is already well-trained while the student has little attribution knowledge. Under this circumstance, the student is likely to fall into a local optimum. To enable smooth knowledge distillation, we normalize the attribution maps before minimizing the difference. Concretely, we first transform the single-view attribution maps into unit vectors:
\begin{equation}
\small
    \widetilde{\mathbf{a}}^{t,c} = \frac{\mathbf{a}^{t,c}}{\Vert \mathbf{a}^{t,c} \Vert_2}, \
    \widetilde{\mathbf{a}}^{s,c} = \frac{\mathbf{a}^{s,c}}{\Vert \mathbf{a}^{s,c} \Vert_2}.
\end{equation}
Then we reformulate the normalized multi-view attribution maps in Eq. \eqref{eq:multi-view atr maps} as:
\begin{equation}
\small
    \widetilde{\mathbf{A}}^{t} = \parallel_{c=1}^{C} \widetilde{\mathbf{a}}^{t,c},\
    \widetilde{\mathbf{A}}^{s} = \parallel_{c=1}^{C} \widetilde{\mathbf{a}}^{s,c}.
\end{equation}
The normalized attribution maps only preserve the information of relative importance among tokens regardless of their absolute importance, which we believe is the crucial knowledge to transfer. Finally, we define the attribution distillation loss as:
\begin{equation}\label{eq:attribution loss}
\small
     \mathcal{L}_{attr} = \Vert \widetilde{\mathbf{A}}^{t} - \widetilde{\mathbf{A}}^{s} \Vert_2.
\end{equation}

\subsubsection{Overall Objective}
%For the overall training objective, we combines the original cross-entropy loss between the output of student and hard label with the logit-based KD loss and our proposed attribution-based KD loss:
We combine the original cross-entropy loss between the output of the student and the ground-truth label, the response-based loss (on the logits) \cite{hinton2015distilling}, and the proposed attribution-driven distillation loss to train the student model. The overall objective is defined as:
\begin{equation}
\small
     \mathcal{L} = (1-\alpha) \mathcal{L}_{ce} + \alpha \mathcal{L}_{logit} + \beta \mathcal{L}_{attr},
\end{equation}
where $\mathcal{L}_{ce}$ = $-\text{log}\sigma(z^s)[y]$ is the cross-entropy loss and $\mathcal{L}_{logit}$=$\text{KL}(\sigma(\frac{z^t}{\tau}) \Vert \sigma(\frac{z^s}{\tau}))$ is the loss on the output logits. And, $\alpha$ and $\beta$ are two hyperparameters, $\sigma$ is the softmax function, $y$ is the ground-truth label, $\tau$ is the temperature, and $z^t$ and $z^s$ are the output logits of the teacher and student models, respectively. $\text{KL}(\cdot)$ denotes the KL-divergence.

\begin{table*}[t]
        
	\begin{adjustbox}{width=0.78\width,center}
	\begin{tabular}{l|c|cccccccc| >{\columncolor{mygray}} c} 
		\toprule
		Model & \#Params & \makecell[c]{CoLA\\(Mcc)} &\makecell[c]{MNLI-(m/mm)\\(Acc)} &\makecell[c]{SST-2\\(Acc)} &\makecell[c]{QNLI\\(Acc)} &\makecell[c]{MRPC\\(F1)} &\makecell[c]{QQP\\(Acc)} &\makecell[c]{RTE\\(Acc)} &\makecell[c]{STS-B\\(Spear)} &Avg\\ 
		\hline
		\multicolumn{11}{c}{\textit{Dev}} \\
            \hline
		$\text{BERT}_{base}\text{ (Teacher)}$ &110M &60.3 &84.9/84.8 &93.7 &91.7 &91.4 &91.5 &69.7 &89.4 &84.1 \\
            $\text{BERT}_{6}\text{ (Student)}$ &66M &51.2 &81.7/82.6 &91.0 &89.3 &89.2 &90.4 &66.1 &88.3 &80.9 \\
            \hline
            Vanilla KD \cite{hinton2015distilling} &66M &53.6 &82.7/83.1 &91.1 &90.1 &89.4 &90.5 &66.8 &88.7 &81.6 \\
            PD \citep{turc2019well} &66M &- &82.5/83.4 &91.1 &89.4 &89.4 &\underline{90.7} &66.7 &- &- \\
            PKD \citep{sun-etal-2019-patient} &66M &45.5 &81.3/- &91.3 &88.4 &85.7 &88.4 &66.5 &86.2 &79.2 \\
            TinyBERT \citep{jiao-etal-2020-tinybert} &66M &53.8 &83.1/83.4 &\underline{92.3} &89.9 &88.8 &90.5 &66.9 &88.3 &81.7 \\
            CKD \citep{park-etal-2021-distilling} &66M &\underline{55.1} &\textbf{83.6}/ \underline{84.1} &\textbf{93.0} &\underline{90.5} &89.6 &\textbf{91.2} &67.3 &\underline{89.0} &\underline{82.4} \\
            MGSKD \citep{liu-etal-2022-multi-granularity} &66M &49.1 &83.3/83.9 &91.7 &90.3 &\underline{89.8} &\textbf{91.2} &\underline{67.9} &88.5 &81.5 \\
            \rowcolor{mygray} AD-KD &66M &\textbf{58.3} &\underline{83.4}/\textbf{84.2} &91.9 &\textbf{91.2} &\textbf{91.2} &\textbf{91.2} &\textbf{70.9} &\textbf{89.2} &\textbf{83.4} \\
            \hline
		\multicolumn{11}{c}{\textit{Test}}\\
            \hline
            $\text{BERT}_{base}\text{ (Teacher)}$ &110M &51.5 &84.5/84.1 &94.1 &90.9 &87.7 &89.2 &67.5 &85.5 &81.4 \\
            $\text{BERT}_{6}\text{ (Student)}$ &66M &41.7 &81.9/81.0 &91.3 &88.9 &85.2 &88.0 &64.0 &82.4 &77.9 \\
            \hline
            Vanilla KD \citep{hinton2015distilling} &66M &42.3 &82.7/81.8 &\underline{92.0} &89.3 &86.3 &88.2 &65.0 &\underline{82.7} &78.6 \\
		PD \citep{turc2019well} &66M &- &82.8/82.2 &91.8 &88.9 &86.8 &\underline{88.9} &65.3 &- &- \\
            PKD \citep{sun-etal-2019-patient} &66M &\underline{43.5} &81.5/81.0 &\underline{92.0} &89.0 &85.0 &\underline{88.9} &\underline{65.5} &81.6 &78.4 \\
            MGSKD \citep{liu-etal-2022-multi-granularity} &66M &42.8 &\textbf{83.4}/\textbf{82.8} &\textbf{92.1} &\underline{89.5} &\underline{87.0} &\textbf{89.1} &63.7 &82.2 &\underline{78.7} \\
            \rowcolor{mygray} AD-KD &66M &\textbf{47.0} &\underline{83.1}/\underline{82.6} &91.8 &\textbf{90.0} &\textbf{87.1} &\underline{88.9} &\textbf{65.8} &\textbf{83.4} &\textbf{79.6} \\
            \bottomrule
	\end{tabular}
        
        \end{adjustbox}
        \caption{Overall results on the GLUE benchmark. The results of baselines except vanilla KD and MGSKD are imported from \citet{park-etal-2021-distilling}. Results of development sets are averaged over 3 runs and we submit the model with the highest score to the official GLUE server to obtain the results of test sets. Average score is computed excluding the MNLI-mm accuracy. The best results of the student models are shown in bold and the second best results are shown with underline. Results are statistically significant with p-value $<$ 0.005.}	
	\label{tab:overall}
 
\end{table*}
\section{Experimental Settings}
\subsection{Datasets}
We evaluate our method on eight tasks of the GLUE benchmark \citep{wang-etal-2018-glue}, including CoLA \citep{warstadt-etal-2019-neural}, MNLI \citep{williams-etal-2018-broad}, SST-2 \citep{socher-etal-2013-recursive}, QNLI \citep{rajpurkar-etal-2016-squad}, MRPC \citep{dolan-brockett-2005-automatically}, QQP \citep{chen2018quora}, RTE \citep{bentivogli2009fifth} and STS-B \citep{cer-etal-2017-semeval}. The details of these datasets are introduced in Appendix \ref{sec:datasets}. For evaluation metrics, we follow previous works \citep{park-etal-2021-distilling,liu-etal-2022-multi-granularity} and report accuracy on MNLI, SST-2, QNLI, QQP and RTE, F1 score on MRPC, Matthews correlation coefficient on CoLA, and Spearman’s rank correlation coefficient on STS-B.

\subsection{Baseline Methods}
%In this paper, we focus on task-specific distillation and do not augment the training sets as \citet{liu-etal-2022-multi-granularity}. We compare our method with logit-based KD baseline PD \citep{turc2019well}, which we also reimplement and denote as Vanilla KD. Other intermediate feature-based KD baselines are also included in our experiments: PKD \citep{sun-etal-2019-patient} which distills the hidden representations, TinyBERT \citep{jiao-etal-2020-tinybert} which distills the self-attention matrices, as well as CKD \citep{park-etal-2021-distilling} and MGSKD \citep{liu-etal-2022-multi-granularity} which distill the relation between hidden representations. For a fair comparison, MiniLM \citep{wang2020minilm,wang-etal-2021-minilmv2} and MobileBERT \citep{sun-etal-2020-mobilebert} are not compared due to their two-stage distillation setting which involves both task-agnostic and task-specific distillation. What's more, MGSKD \citep{liu-etal-2022-multi-granularity} adopts augmented training data for task-specific distillation and uses a different teacher-student pair. Their method also depends on an external classifier-based English chunker to extract the span-level information. Therefore, we reimplement MGSKD under our setting and ablate the span-level relation.

%In this paper, we focus on task-specific distillation and do not augment the training sets as \citet{liu-etal-2022-multi-granularity}. 
We compare AD-KD with response-based KD methods and several state-of-the-art feature-based and relation-based KD methods. Response-based baselines include Vanilla KD \cite{hinton2015distilling} and PD \cite{turc2019well}. Feature-based and relation-based baselines include PKD \cite{sun-etal-2019-patient} which distills the hidden representations, TinyBERT \cite{jiao-etal-2020-tinybert} which distills the self-attention matrices, and CKD \citep{park-etal-2021-distilling} and MGSKD \citep{liu-etal-2022-multi-granularity} which distill the relation between hidden representations. For a fair comparison, MiniLM \citep{wang2020minilm,wang-etal-2021-minilmv2} and MobileBERT \citep{sun-etal-2020-mobilebert} are not presented due to their two-stage distillation settings which involve both task-agnostic and task-specific distillation. Our AD-KD focuses on task-specific distillation and does not augment the training sets. Moreover, MGSKD \citep{liu-etal-2022-multi-granularity} only reports results on a 4-layer BERT student model which is different from other baselines. To ensure a fair comparison, we re-implemented MGSKD using their released code to obtain a 6-layer student model. The original MGSKD approach also relies on span-level information that is extracted from external knowledge sources, which is not publicly available nor included in other baselines. Therefore, we did not use this external knowledge in our re-implementation of MGSKD. 

\subsection{Implementation Details}
Our code is implemented in Pytorch with the Transformers package \citep{wolf-etal-2020-transformers}. We fine-tune $\text{BERT}_{base}$ as the teacher model, and utilize a smaller BERT released by \citet{turc2019well} with 6 Transformer layers, 768 hidden neurons and 12 attention heads to instantiate the student model following \citet{park-etal-2021-distilling}. We search for the optimal learning rate in \{2e-5, 3e-5, 4e-5, 5e-5\}, $\alpha$ in \{0.8, 0.9, 1.0\} and temperature $\tau$ in \{1, 2, 3, 4\}. For the hyperparameter $\beta$, we tune within \{1, 10, 50, 100\}. For the IG steps $m$ described in Section \ref{sec:background}, we adopt $m$ = 1 in the main results due to the huge computational overhead. Part of results with $m$ varying from 1 to 8 are reported in Section \ref{sec:ig steps}.
%Since experimenting with all possible $K$ results in a large search space, we first set K=1 and search for the above hyperparameters. Then the optimal combination of these hyperparameters are kept fixed and 
$K$ is empirically searched within \{384, 512, 640, 700, 734, 768\}. Results with different values of $K$ are also reported. The detailed hyperparameter settings and training cost are provided in Appendix \ref{sec:hyper}. Our code is available at \url{https://github.com/brucewsy/AD-KD}.

\section{Results and Analysis}
\subsection{Main Results}
The main results are presented in Table \ref{tab:overall}. It can be seen that AD-KD outperforms all baselines on most of the datasets. Specifically, AD-KD yields an average improvement of 1.0 and 1.9 points over CKD and MGSKD respectively on development sets, and another average improvement of 0.9 points over MGSKD on test sets. Note that other feature-based and relation-based KD methods even under-perform vanilla KD, indicating the difficulty of aligning the teacher and the student at intermediate layers. In contrast, AD-KD distills the attribution knowledge from a global perspective which is more data-specific and shows significant improvement over vanilla KD. We provide two cases in Appendix \ref{sec:case study} to intuitively demonstrate the strength of AD-KD. We also observe that AD-KD does not show a satisfying performance on SST-2. We believe the reason is that the sentences in SST-2 are much shorter than those in other datasets, and in this case, the student is likely to already capture the attribution knowledge implicitly from the soft-labels of the teacher \citep{zhang2022quantifying}.

\subsection{Ablation Study}

\begin{table*}[t]
        
        \begin{adjustbox}{width=0.82\width,center}
	\begin{tabular}{l|cccccccc} 
		\toprule
		Method & \makecell[c]{CoLA\\(Mcc)} &\makecell[c]{MNLI-(m/mm)\\(Acc)} &\makecell[c]{SST-2\\(Acc)} &\makecell[c]{QNLI\\(Acc)} &\makecell[c]{MRPC\\(F1)} &\makecell[c]{QQP\\(Acc)} &\makecell[c]{RTE\\(Acc)} &\makecell[c]{STS-B\\(Spear)}\\ 
		\hline
            AD-KD &58.3 &83.4/84.2 &91.9 &91.2 &91.2 &91.2 &70.9 &89.2  \\
            \ \ w/o $\mathcal{L}_{attr}$ &53.6 &82.7/83.1 &91.2 &90.2 &89.2 &90.5 &67.5 &88.9  \\
            %\ \ with shuffled $\mathcal{L}_{attr}$ &51.3 &82.9/83.5 &91.4 &89.5 &88.4 &90.6 &67.5 &88.4  \\
            \ \ w/o $\mathcal{L}_{ce}$ &57.8 &83.6/84.1 &91.3 &90.8 &90.8 &91.2 &69.3 &88.9 \\
            \ \ w/o $\mathcal{L}_{logit}$ &53.9 &81.9/82.8 &91.1 &90.5 &89.9 &90.9 &68.6 &88.8 \\
            \bottomrule
	\end{tabular}
        \end{adjustbox}
        \caption{Ablation study of different loss terms. The results are based on GLUE development sets.}
	\label{tab:ablation of AM}
 \vspace{-4mm}
\end{table*}

%\noindent \textbf{Types of attribution maps.} ~To analyze the impact of different types of attributions maps, we conduct ablation experiments on two variants of AD-KD: AD-KD without attribution maps (i.e., vanilla KD) and AD-KD with shuffled attribution maps of teacher. As reported in Table \ref{tab:ablation of AM}, again we see an obvious drop after removing the distillation of attribution maps. Moreover, distilling the shuffled attribution maps of teacher generally performs worst, which indicates that the order of attribution maps is crucial since it contains the correct information about relative importance of different tokens.

\noindent \textbf{Impact of Loss Terms} ~To analyze the impact of different loss terms, we conduct ablation experiments on three variants of AD-KD: (1) AD-KD without attribution distillation (i.e., vanilla KD), (2) AD-KD without the original cross-entropy loss, and (3) AD-KD without logit distillation. As reported in Table~\ref{tab:ablation of AM}, again we observe an obvious performance drop after removing the attribution distillation. We also note that removing either the conventional cross-entropy loss or logit distillation loss causes noticeable performance degradation, suggesting both of them contribute to the improvement of AD-KD. Nevertheless, our attribution distillation contributes most to the performance of AD-KD, showing that data-specific reasoning information is crucial in knowledge distillation. 

\begin{figure}[t]
	\centering
        \includegraphics[scale=0.53]{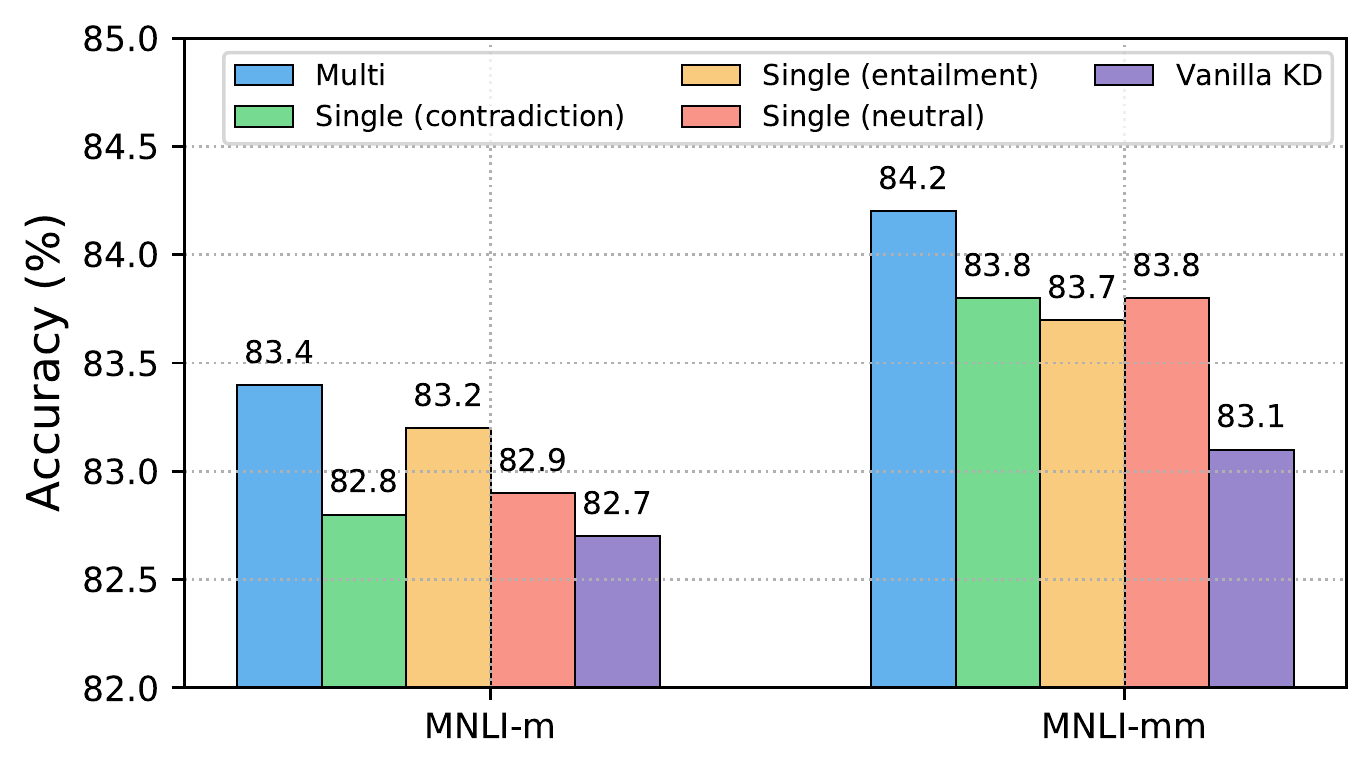}
	\caption{Ablation study of multi-view attribution on the MNLI development set.}
	\label{fig:ablation mnli}
  \vspace{-4mm}
\end{figure}

\noindent \textbf{Multi-view Attribution} ~In AD-KD, the student learns the attribution knowledge from a variety of possible outputs to get a better understanding of the teacher. Here we study how the number of attribution views affects the final results. Experiments are conducted on MNLI which is a multi-classification task including three labels: entailment, contradiction, and neutral. We make a comparison between multi-view attribution and single-view attribution w.r.t. each candidate label respectively. The results are shown in Figure \ref{fig:ablation mnli}, from which we note that each of the single-view attributions plays a positive role and is superior to vanilla KD. Moreover, combining all attribution views yields further performance improvement, demonstrating that multi-view attribution is more preferable for distillation.

\begin{figure}[t]
	\centering
        \includegraphics[scale=0.56]{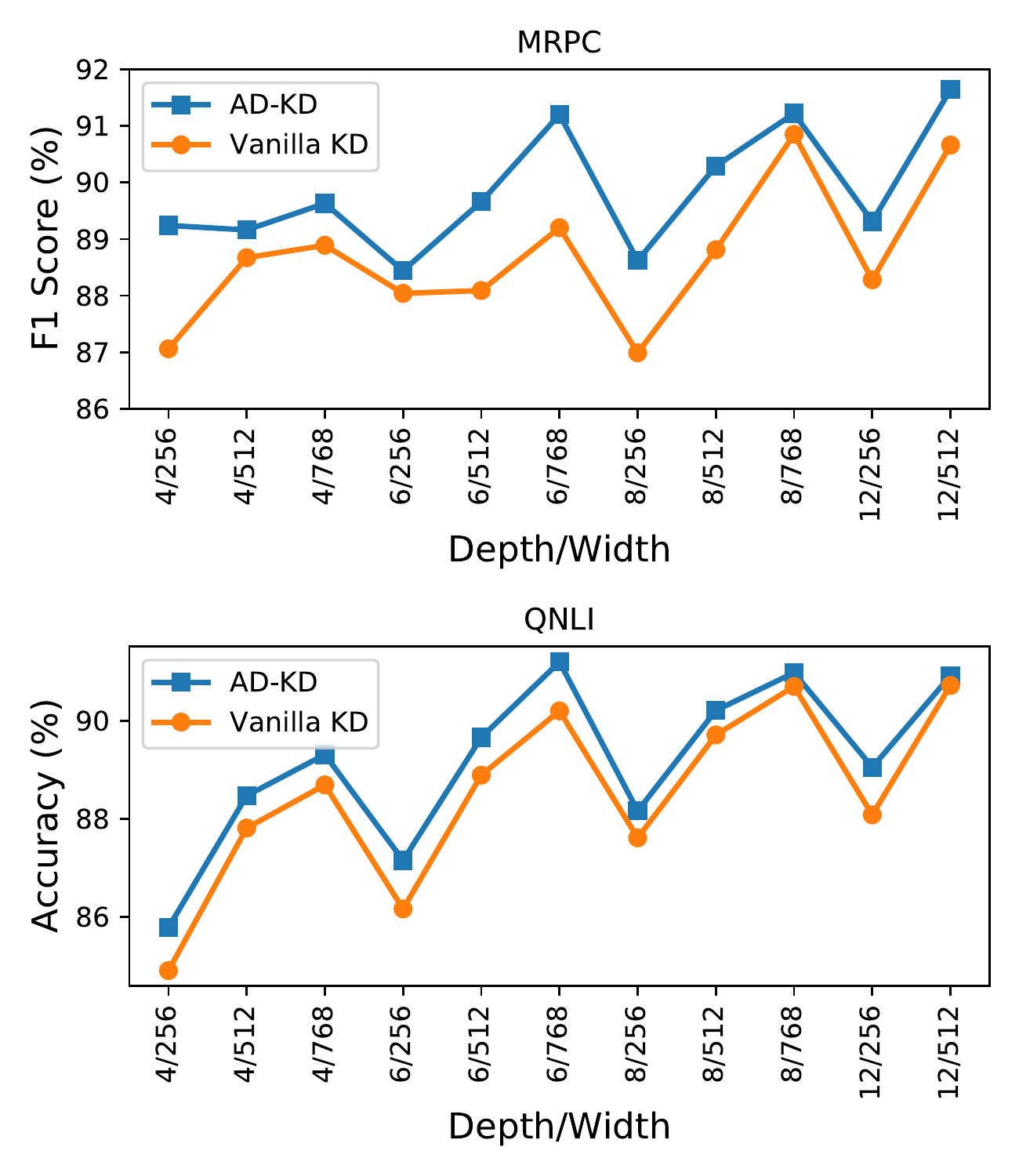}
	\caption{Results of AD-KD and vanilla KD on MRPC and QNLI development sets at different student scales.}
	\label{fig:architectue}
 \vspace{-4mm}
\end{figure}

\noindent \textbf{Student Model Size} ~To investigate whether AD-KD can boost the performance across different sizes of student, we further compare AD-KD with vanilla KD on MRPC and QNLI under various student scales provided by \citet{turc2019well}. As observed in Figure \ref{fig:architectue}, AD-KD consistently outperforms vanilla KD, which validates the effectiveness and stability of our approach.

%\begin{figure}[t]
	%\centering
        %\includegraphics[scale=0.57]{figures/architecture_mrpc.pdf}
	%\caption{Results of AD-KD and vanilla KD on MRPC development set as the size of student changes.}
	%\label{fig:architectue mrpc}
%\end{figure}

%\begin{figure}[t]
	%\centering
        %\includegraphics[scale=0.57]{figures/architecture_qnli.pdf}
	%\caption{Results of AD-KD and vanilla KD on QNLI development set as the size of student changes.}
	%\label{fig:architectue qnli}
%\end{figure}

\subsection{Impact of Top-\textit{K}}
\begin{figure}
	\centering
        \includegraphics[scale=0.53]{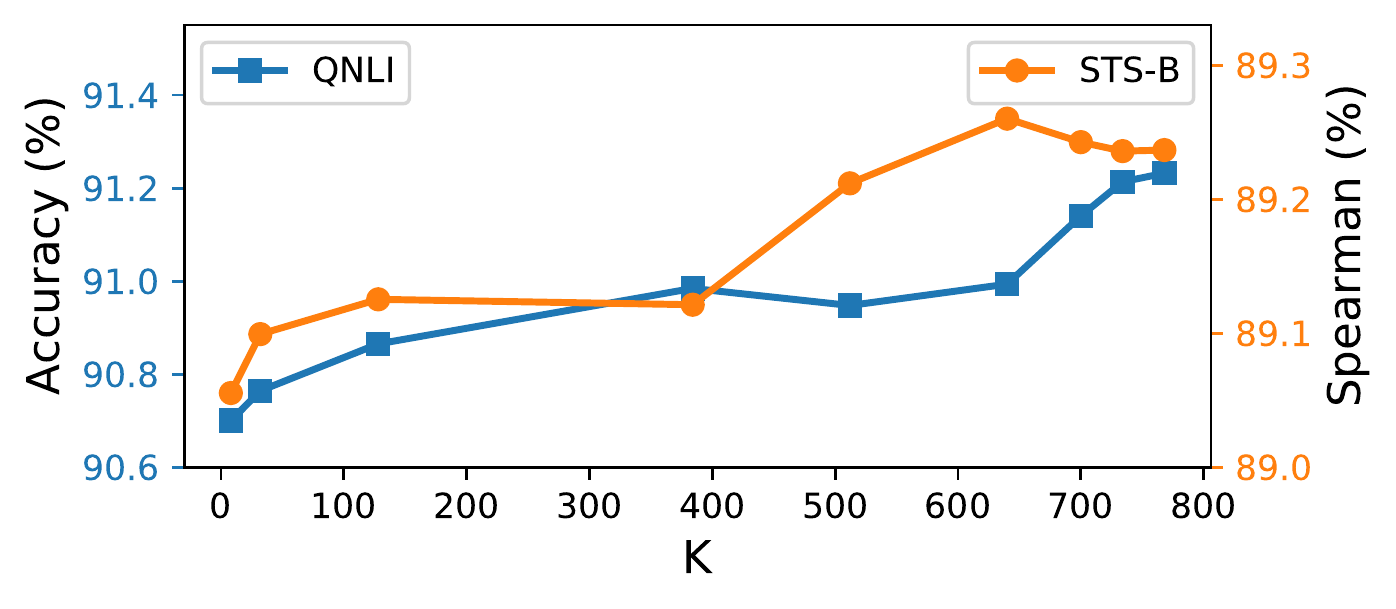}
	\caption{Results on STS-B and QNLI development sets as the number ($K$) of retained dimensions changes.}
	\label{fig:topk}
 \vspace{-4mm}
\end{figure}

Recall that in order to eliminate the interference of noisy dimension, AD-KD adopts the top-\textit{K} approach on the input embeddings of the teacher to filter out the dimensions with relatively low attribution scores. In this section, we conduct in-depth analysis on the impact of $K$. We conduct experiments on STS-B and QNLI, and plot the results with different values of $K$ in Figure \ref{fig:topk}. As illustrated in the figure, the performance on the small dataset STS-B (7k) first improves as $K$ increases and then slightly degrades after $K$ exceeds 600. However, the performance on the larger dataset QNLI (108k) improves almost monotonically with the increasing of $K$. We conjecture that choosing a suitable $K$ is beneficial on small datasets since there are probably more noisy dimensions in the input embeddings of the teacher, while preserving all dimensions may be preferable on larger datasets.

\subsection{Impact of IG Steps} \label{sec:ig steps}
\begin{figure}
	\centering
        \includegraphics[scale=0.53]{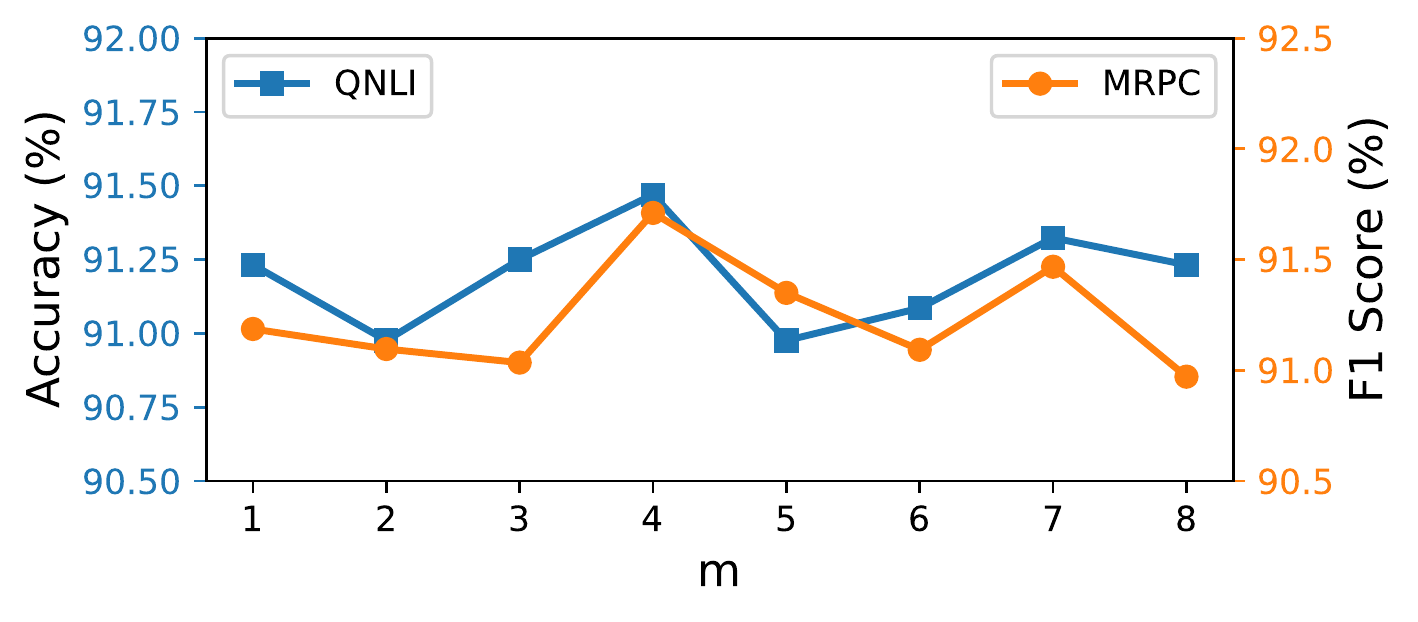}
	\caption{Results on MRPC and QNLI development sets as the number ($m$) of IG steps changes.}
	\label{fig:ig steps}
 %\vspace{-4mm}
\end{figure}

In our experiments, the IG steps $m$ are set to 1 by default when extracting the attribution maps. In this section, we provide more results with different values of $m$ in Figure \ref{fig:ig steps} to understand its impact on distillation. We observe that as $m$ increases, the performance of AD-KD fluctuates in a certain range. Although it is possible to find a point that surpasses our default setting and even the teacher, identifying the optimal value of $m$ for each task is costly since a large $m$ causes huge computational overhead. In contrast, $m$=1 achieves better trade-off between performance and computational cost.

\begin{table}[!h]
	
 	\begin{adjustbox}{width=0.84\width,center}
	\begin{tabular}{l|ll}
		\toprule
		Attribution Layer & \makecell[c]{MRPC\\(F1)} & \makecell[c]{QNLI\\(Acc)} \\
		\hline
		input &\textbf{91.2} &\textbf{91.2} \\
		first &90.5 &90.9  \\ 
		penultimate  &90.4  &90.9  \\ 
            uniform &90.6 &91.1 \\
            input \& uniform &90.1 &90.6 \\
		\bottomrule
	\end{tabular}
 	\end{adjustbox}
        \caption{Results of different attribution layers on MRPC and QNLI development sets.}
	\label{tab:attribution layer}
 \vspace{-5mm}
\end{table}

\subsection{Attribution Distillation Layer}
Apart from the attribution knowledge of input layer, the attribution knowledge of intermediate layers can also be transferred during distillation. To confirm the motivation that the former is better than the latter, we conduct experiments on MRPC and QNLI with different attribution layers. Specifically, we choose the first layer and the penultimate layer for comparison. Besides, we also try a uniform strategy which is widely adopted as the mapping function between the teacher and the student layers \cite{jiao-etal-2020-tinybert,park-etal-2021-distilling,liu-etal-2022-multi-granularity}. From the results shown in Table \ref{tab:attribution layer}, we see that uniform mapping strategy performs best among intermediate layer methods. However, neither of these intermediate layers outperforms input layer, indicating that the attribution knowledge of intermediate layers is more model-specific and difficult to transfer. In addition, distilling the knowledge jointly from the input and the intermediate layers does not improve the performance.

\subsection{Impact of $\alpha$ and $\beta$}
\begin{figure}[t]
	\centering
        \includegraphics[scale=0.50]{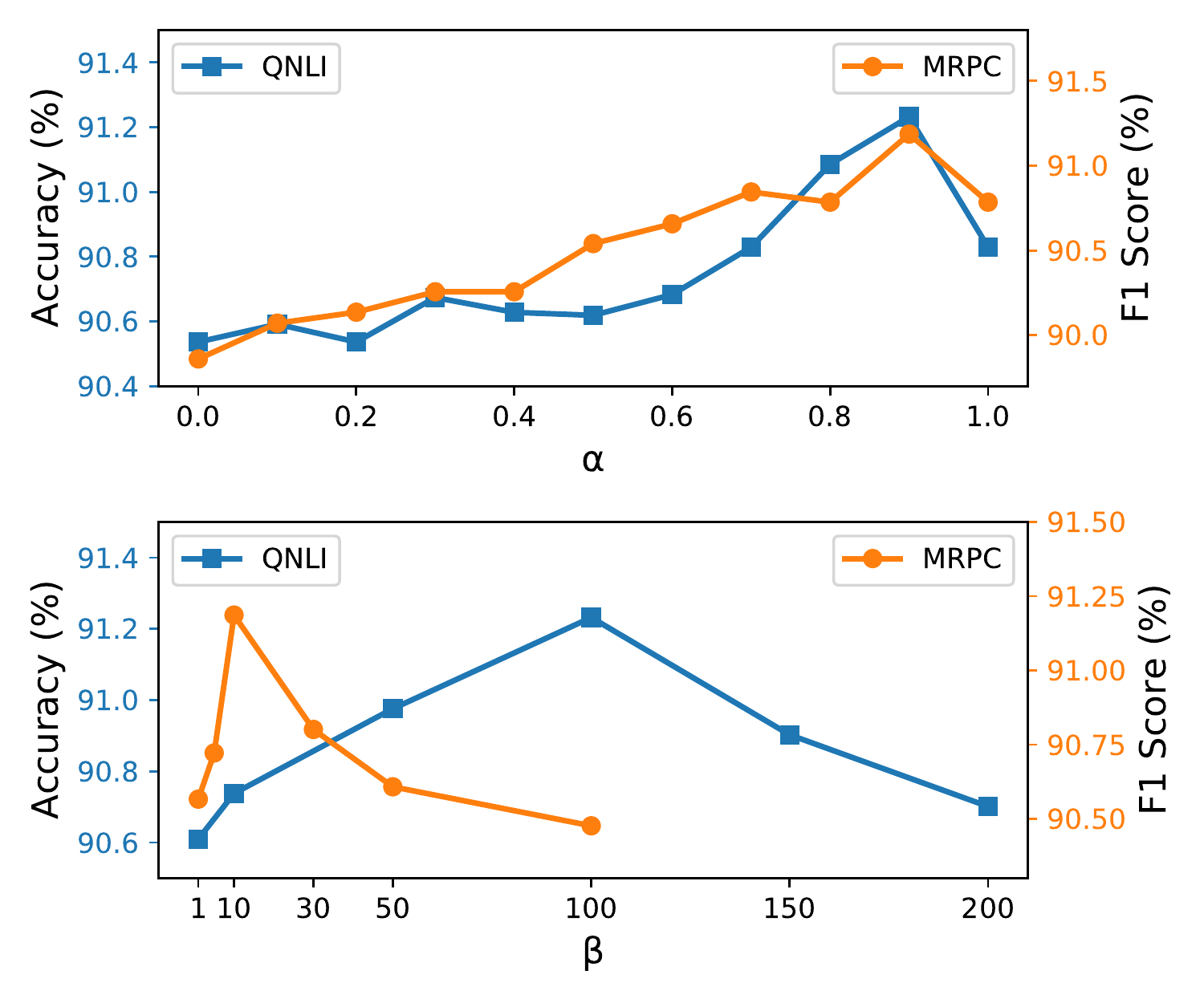}
	\caption{Results on MRPC and QNLI development sets as $\alpha$ and $\beta$ changes.}
	\label{fig:alpha_beta}
 \vspace{-5mm}
\end{figure}

For the training objective of AD-KD, we introduce $\alpha$ and $\beta$ to balance the original cross-entropy loss, logit distillation loss, and attribution distillation loss. To investigate their impact on model performance, we show the results of different values of $\alpha$ and $\beta$ on MRPC and QNLI in Figure \ref{fig:alpha_beta}, where we fix one while altering the other. We observe a unified trend across different tasks that when $\alpha$ is small, the student does not perform well due to the lack of response-based knowledge of the teacher, and when $\alpha$ is around 0.9, the student performs best. Therefore, we select $\alpha$ close to 1. We also observe from the figure that as $\beta$ increases, the performance first keeps improving and reaches the peak, then it starts to decline. Unlike $\alpha$, however, the optimal value of $\beta$ varies with different tasks, indicating that $\beta$ is more sensitive to the task compared to $\alpha$. More discussion of $\beta$ are given in Appendix \ref{sec:more results of beta}.

%Also, from Figure \ref{fig:beta mrpc} and Figure \ref{fig:beta qnli} we can make other two observations. For one thing, by altering $\beta$, the tendency of attribution gap on development sets is consistent with the one on training sets, which indicates that the attribution knowledge learned from training data can be well generalized to unseen data. For another, however, a smaller gap of attribution maps between teacher and student does not always lead to better performance, which means we still need to meticulously tune $\beta$ for different tasks and data.

%\begin{figure}[b]
	%\centering
        %\includegraphics[scale=0.55]{figures/alpha_experiment.pdf}
	%\caption{Results on MRPC and QNLI development sets as $\alpha$ changes.}
	%\label{fig:alpha}
%\end{figure}

%\begin{figure}[b]
	%\centering
        %\includegraphics[scale=0.55]{figures/beta_experiment_mrpc.pdf}
	%\caption{Results on MRPC development set as $\beta$ changes.}
	%\label{fig:beta mrpc}
%\end{figure}

%\begin{figure}[b]
	%\centering
        %\includegraphics[scale=0.55]{figures/beta_experiment_qnli.pdf}
	%\caption{Results on QNLI development set as $\beta$ changes.}
	%\label{fig:beta qnli}
%\end{figure}

%\vspace{-2mm}
\section{Conclusion}
In this paper, we propose AD-KD, a novel knowledge distillation framework for language model compression. Unlike other distillation methods, AD-KD investigates the model knowledge from the perspective of input attribution, which is vital yet easy to transfer between the teacher and the student. Moreover, top-\textit{K} method is adopted to obtain noiseless attribution maps among input tokens, and multi-view attribution is conducted for a more comprehensive distillation. To our knowledge, this is the first work that incorporates attribution into knowledge distillation. Extensive experiments including ablation studies are carried out to show the effectiveness of AD-KD and its components. 
%Furthermore, we also show that top-\textit{K} is preferable for smaller datasets and distilling the attribution knowledge of the input layer achieves the best performance.
With the recent emergence of large language models (LLMs), gradient-based attribution methods are infeasible due to the unavailable parameters. However, the idea of AD-KD can still be potentially extended to these black-box models by using occlusion-based attribution or using chain-of-thoughts \citep{wei2022chain} as the rationale for distillation. We will leave it to future work.

\section*{Acknowledgements}
This work was supported by the National Natural Science Foundation of China (No. 62176270), the Guangdong Basic and Applied Basic Research Foundation (No. 2023A1515012832), and the Program for Guangdong Introducing Innovative and Entrepreneurial Teams (No. 2017ZT07X355).

\section*{Limitations}
This work introduces the general idea of incorporating attribution into knowledge distillation, and there are three potential limitations. First, although AD-KD chooses Integrated Gradients for attribution, there are actually other attribution methods \citep{janizek2021explaining,sikdar-etal-2021-integrated} which can also be fitted in our framework. The question of whether these methods perform better than Integrated Gradients when combined with knowledge distillation is still unclear. Second, we conduct experiments on BERT of different scales and have not yet validated the effectiveness of AD-KD on other model structures. Third, while we only perform task-specific knowledge distillation in our experiments, applying AD-KD to task-agnostic knowledge distillation is also worth investigating.

\section*{Ethics Statement}
Our work will not cause ethical issues and the datasets used in this paper are publicly available.

\bibliography{anthology,custom}
\bibliographystyle{acl_natbib}

\clearpage
\appendix

\section{Experimental Details}
\subsection{Details of Datasets}\label{sec:datasets}
We evaluate AD-KD on eight tasks of GLUE benchmark \citep{wang-etal-2018-glue}. Specifically, there are two single-sentence tasks: CoLA \citep{warstadt-etal-2019-neural} which aims to predict if the given sentence is grammatically correct, and SST-2 \citep{socher-etal-2013-recursive} which aims to predict the sentiment of the given sentence; two paraphrase tasks: MRPC \citep{dolan-brockett-2005-automatically} which aims to predict if two given sentences are semantically equivalent, and QQP \citep{chen2018quora} which is similar to MRPC; three inference tasks which aim to predict if the premise entails the hypothesis: MNLI \citep{williams-etal-2018-broad}, QNLI \citep{rajpurkar-etal-2016-squad}, and RTE \citep{bentivogli2009fifth}; and one similarity task: STS-B \citep{cer-etal-2017-semeval} which aims to predict a continual score measuring the semantic similarity between a pair of sentences. The statistics of these datasets are shown in Table \ref{tab:dataset}.

\begin{table}[h]
        \begin{adjustbox}{width=0.8\width,center}
	\begin{tabular}{l|cccc} 
		\toprule
		Task &\#Train &\#Dev &\#Test &\#Label \\
            \hline
            \multicolumn{5}{c}{Single-Sentence Classification} \\
		\hline
            CoLA &8.5k &1k &1k &2 \\
            SST-2 &67k &872 &1.8k &2 \\
            \hline
            \multicolumn{5}{c}{Pairwise Text Classification} \\
            \hline
            MNLI &393k &20k &20k &3 \\
            QNLI &108k &5.7k &5.7k &2 \\
            MRPC &3.7k &408 &1.7k &2 \\
            QQP &364k &40k &391k &2 \\
            RTE &2.5k &276 &3k &2 \\
            \hline
            \multicolumn{5}{c}{Text Similarity} \\
            \hline
            STS-B &7k &1.5k &1.4k &1 \\
            \bottomrule
	\end{tabular}
        \end{adjustbox}
        \caption{Statistics of the GLUE datasets.}
	\label{tab:dataset}
 \vspace{-5mm}
\end{table}

\subsection{Hyperparameter Settings}\label{sec:hyper}
We run all experiments on GeForce RTX 2080 Ti GPUs. Table \ref{tab:hyperparameter} presents the hyperparameter settings and training costs of AD-KD on GLUE tasks. Generally, AD-KD runs 1.2 to 3 times slower compared to vanilla KD on different tasks, due to the extra back-propagation. However, all students obtained by different distillation methods have the same inference speed.

\section{More Discussion} \label{sec:discussion}
In this section, we discuss the difference between distilling the attribution maps and distilling the attention matrices. In a sense, attention matrices are similar to attribution maps since they both reflect the contribution that each input token makes on a model prediction to some extent \citep{bastings-filippova-2020-elephant,xu-etal-2020-self}. However, there are several drawbacks when it comes to distillation. On one hand, attention correlates well with attribution locally in specific layers and heads but not globally, indicating that attention maps are inadequate to draw conclusions that refer to the input of the model \citep{pascual-etal-2021-telling}. In other words, attention matrices are more like model-specific knowledge that are probably challenging for the student to learn due to the layer mapping issue, especially when the student has much fewer parameters than the teacher. On the other hand, some works point out that by adversarial training, alternative attention weights can be found whereas the prediction remains almost the same \citep{jain-wallace-2019-attention,wiegreffe-pinter-2019-attention}. Therefore, an optimal student unnecessarily shares similar attention matrices with its teacher. Our proposed AD-KD adopts a more reliable gradient-based method to obtain the attribution maps, which is shown better than attention matrices employed by baselines.

\begin{table}
\vspace{-3mm}
        \begin{adjustbox}{width=0.55\width,center}
	\begin{tabular}{l|cccccccc} 
		\toprule
		Hyperparameter &CoLA &MNLI &SST-2 &QNLI &MRPC &QQP &RTE &STS-B \\
            \hline 
             \makecell[c]{Learning Rate} &4e-5 &4e-5 &5e-5 &4e-5 &3e-5 &4e-5 &2e-5 &5e-5 \\
             \makecell[c]{Total Batch Size} &32 &64 &32 &32 &16 &32 &16 &16 \\
             \makecell[c]{Max Seq. Length} &128 &128 &128 &128 &128 &128 &128 &128 \\
             \makecell[c]{$\alpha$}  &0.9 &0.8 &0.8 &0.9 &0.9 &1.0 &0.9 &0.8 \\
             \makecell[c]{$\beta$} &1 &10 &1 &100 &10 &50 &10 &1 \\
             \makecell[c]{$\tau$} &1 &3 &2 &3 &4 &4 &2 &3 \\
             \makecell[c]{K} &768 &768 &768 &768 &768 &734 &700 &640 \\
             \makecell[c]{$m$} &1 &1 &1 &1 &1 &1 &1 &1 \\
             \hline
             \makecell[c]{\# GPU} &1 &4 &1 &1 &1 &1 &1 &1 \\
             \makecell[c]{Training Time} &30min &12hr &2.5hr &3hr &20min &16hr &12min &20min \\
            \bottomrule
	\end{tabular}
        \end{adjustbox}
        \caption{Hyperparameter settings and training cost.}
	\label{tab:hyperparameter}
 \vspace{-2mm}
\end{table}

\begin{table}[b]
        \begin{adjustbox}{width=0.8\width,center}
	\begin{tabular}{l|c|c} 
		\toprule
		Model &\#Params &Acc \\
            \hline
            $\text{BERT}_{base}\text{ (Teacher)}$ &110M &68.53 \\
            %$\text{BERT}_{6}\text{ (Student)}$ &66M & \\

		\hline
            Vanilla KD &66M &67.70  \\
            MGSKD &66M &68.29 \\
            AD-KD &66M &68.67 \\
            \bottomrule
	\end{tabular}
        \end{adjustbox}
        \caption{Results on MultiRC development set.}
	\label{tab:multirc}
 \vspace{-5mm}
\end{table}

\section{More Experimental Results}
\subsection{Results on MultiRC}
Considering that the text in GLUE is relatively short (with Max\_Seq\_Length set to 128), We conduct additional experiments on SuperGLUE \citep{wang2019superglue} for more comprehensive evaluation. We select a challenging QA task, MultiRC \citep{khashabi-etal-2018-looking}, with much longer text (with Max\_Seq\_Length set to 512) which requires more attribution knowledge. As shown in Table \ref{tab:multirc}, AD-KD improves 0.97\% over vanilla KD and 0.38\% over MGSKD. Moreover, the performance of AD-KD is on par with the teacher.

 \begin{figure*}[t]
	\centering
        \includegraphics[scale=0.5]{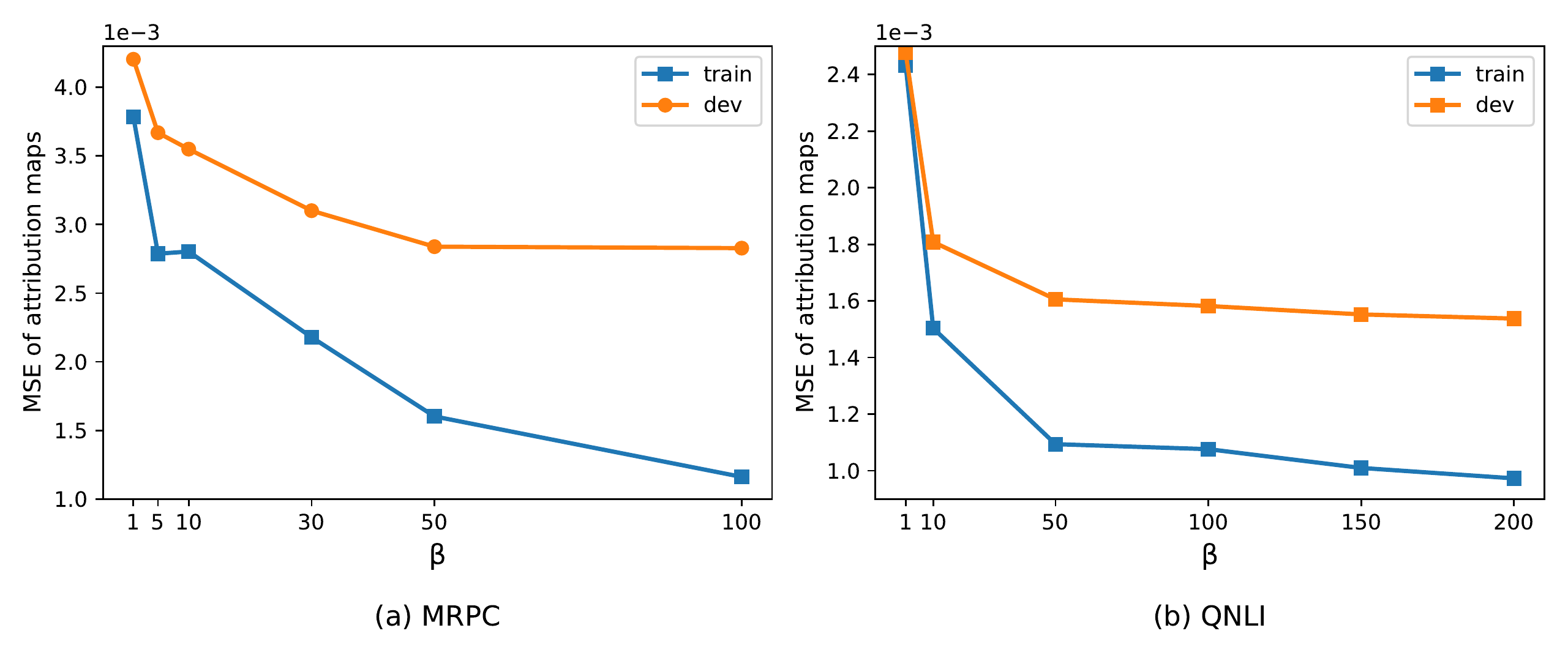}
	\caption{Comparison of the attribution gap between teacher and student on training set and development set.}
	\label{fig:attribution gap}
\end{figure*}

\subsection{Overfitting Study}\label{sec:more results of beta}
In this section, we investigate whether the overfitting problem would happen in attribution distillation. Using Eq. \eqref{eq:attribution loss}, we calculate the attribution gap between the teacher and the student models on the training and development sets of MRPC and QNLI respectively, and show the results in Figure~\ref{fig:attribution gap}. By altering $\beta$, the tendency of attribution gap on development sets is consistent with the one on training sets, which indicates that the attribution knowledge learned from training data can be well generalized to unseen data. Therefore, overfitting tends not to happen in attribution distillation.

 %\begin{figure}[b]
	%\centering
        %\includegraphics[scale=0.62]{figures/attribution_gap_mrpc.pdf}
	%\caption{Comparison of the attribution gap between teacher and student on training set and development set.}
	%\label{fig:attribution gap mrpc}
%\end{figure}
 %\begin{figure}[b]
	%\centering
        %\includegraphics[scale=0.62]{figures/attribution_gap_qnli.pdf}
	%\caption{Comparison of the attribution gap between teacher and student on training set and development set.}
	%\label{fig:attribution gap qnli}
%\end{figure}

\subsection{Case Study}\label{sec:case study}
\begin{figure*}[!ht]
	\centering
        \includegraphics[scale=0.45]{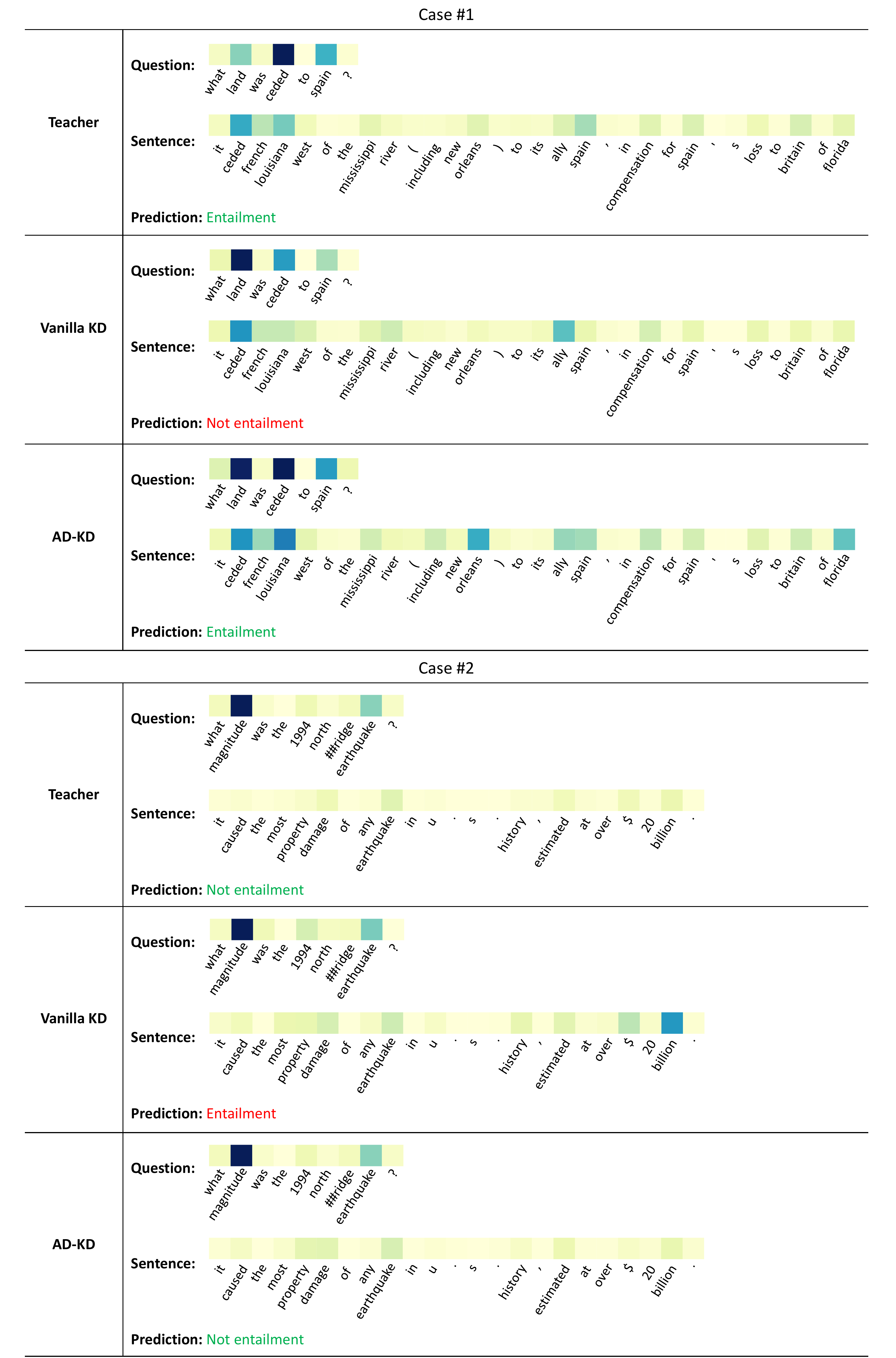}
	\caption{Two illustrative examples of attribution maps and predictions by teacher, vanilla KD and AD-KD from the QNLI development set, where darker colors mean larger attribution scores. In case \#1, AD-KD learns which tokens to focus on (\emph{Louisiana}), while in case \#2, AD-KD learns which tokens not to focus on (\emph{billion}).}
	\label{fig:case}
\end{figure*}

In this section, we provide two examples to show how AD-KD facilitates the imitation of the teacher's reasoning and outperforms vanilla KD. As shown in Figure \ref{fig:case}, vanilla KD makes mistakes by ignoring keyword \emph{Louisiana} or emphasizing an irrelevant word \emph{billion}. In contrast, the attribution maps of AD-KD are more consistent with the ones in the teacher. AD-KD learns what to and not to focus on and thus predicts the label correctly.
\clearpage

\end{document}